\def\year{2018}\relax
\documentclass[letterpaper]{article} %
\usepackage{aaai18}  %
\usepackage{times}  %
\usepackage{helvet}  %
\usepackage{courier}  %
\usepackage{url}  %
\usepackage{graphicx}  
\usepackage[utf8]{inputenc}
\usepackage{textcomp}
\usepackage[table,dvipsnames]{xcolor}
\usepackage{epsfig}
\usepackage{pgfplotstable}
\usepackage{pgfplots}
\usepgfplotslibrary{groupplots}
\usepackage{amsmath}
\usepackage{amssymb}
\usepackage{bbm}
\usepackage{float}
\usepackage[draft]{hyperref}
\usepackage{microtype}
\usepackage{tikz}
\usetikzlibrary{shapes,arrows,arrows.meta, positioning,shapes.misc,decorations.markings}
\usepackage{booktabs}
\usepackage{multirow}
\usepackage{adjustbox}
\usepackage{verbatim}
\usepackage[T1]{fontenc}
\usepackage[title]{appendix}
\usepackage[font={small,singlespacing},labelfont=it,labelsep=period]{caption}
\usepackage{adjustbox} %
\usepackage{graphicx}
\usepackage{caption}
\usepackage{subcaption}
\usepackage{xspace}
\usepackage[noend]{algpseudocode}
\usepackage{algorithm}
\usepackage[defaultlines=2,all]{nowidow}
\usepackage[colorinlistoftodos,textsize=footnotesize]{todonotes} %
\newcommand{\citet}[1]{\citeauthor{#1} ~\shortcite{#1}}
\newcommand{\citep}{\cite}

\newcommand{\textcite}[1]{\citet{#1}}

\usepackage[autostyle, english=american]{csquotes}
\MakeOuterQuote{"}

\makeatletter
\pgfplotsset{
    every axis x label/.append style={
        alias=current axis xlabel
    },
    legend pos/outer south/.style={
        /pgfplots/legend style={
            at={%
                (%
                \@ifundefined{pgf@sh@ns@current axis xlabel}%
                {xticklabel cs:0.5}%
                {current axis xlabel.south}%
                )%
            },
            anchor=north
        }
    }
}
\makeatother

\newcolumntype{t}{>{\ttfamily}l}
\newcolumntype{T}{>{\ttfamily}c}

\newcolumntype{$}{>{\global\let\currentrowstyle\relax}}
\newcolumntype{^}{>{\currentrowstyle}}

\DeclareMathOperator*{\argmax}{arg\,max}
\DeclareMathOperator*{\argmin}{arg\,min}

\algnewcommand\algorithmicforeach{\textbf{for each}}
\algdef{S}[FOR]{ForEach}[1]{\algorithmicforeach\ #1\ \algorithmicdo}

\def\dtf{$\text{DT}^F$\xspace}
\def\dtfc{$\text{DT}^F_c$\xspace}
\def\rff{$\text{RF}^F$\xspace}
\def\rffc{$\text{RF}^F_c$\xspace}
\def\lrf{$\text{LR}^F$\xspace}
\def\nbf{$\text{NB}^F$\xspace}

\begin{document}

\everypar{\looseness=-1 } %
\linepenalty=1000 %

\title{Fair Forests: Regularized Tree Induction to Minimize Model Bias}

\author{Edward Raff \\Booz Allen Hamilton, Strategic Innovation Group \\ University of Maryland, Baltimore County
\And Jared Sylvester  \and Steven Mills \\ Booz Allen Hamilton, Strategic Innovation Group}

\maketitle

\begin{abstract}
The potential lack of fairness in the outputs of machine learning algorithms has recently gained attention both within the research community as well as in society more broadly.
Surprisingly, there is no prior work developing tree-induction algorithms for building fair decision trees or fair random forests. These methods have widespread popularity as they are one of the few to be simultaneously interpretable, non-linear, and easy-to-use. 
In this paper we develop, to our knowledge, the first technique for the induction of fair decision trees.
We show that our "Fair Forest" retains the benefits of the tree-based approach, while providing both greater accuracy and fairness than other alternatives, for both ``group fairness'' and ``individual fairness.''
We also introduce new measures for fairness which are able to handle multinomial and continues attributes as well as regression problems, as opposed to binary attributes and labels only.
Finally, we demonstrate a new, more robust evaluation procedure for algorithms that considers the dataset in its entirety rather than only a specific protected attribute.
\end{abstract}

\section{Introduction}

As applications of Machine Learning becomes more pervasive in society, it is important to consider the fairness of such models. We consider a model to be fair with respect to some protected attribute $a_p$ (such as age or gender), if it's predicted label $\hat{y}$ with respect to a datumn $x$ is unaffected by changes to $a_p$. Removing $a_p$ from $x$ is not sufficient to meet this goal in practice, as $a_p$'s effect is still present as a latent variable \cite{Pedreshi:2008:DDM:1401890.1401959}.
In this work, we look at adapting decision trees, specifically Random Forests, to this problem. Given an attribute $a_p$ that we wish to protect, we will show how to induce a "Fair Forest" that provides improved fairness and accuracy compared to existing approaches.  

Decision Trees have become one of the most widely used classes of machine learning algorithms. In particular, C4.5~\cite{Quinlan1993} and \textsc{cart}~\cite{Breiman1984} tree induction approaches, combined with ensembling approaches like Random Forests~\cite{Breiman2001} and Gradient Boosting~\cite{Friedman2002}, have proven to be potent and effective across a broad spectrum of needs and tasks. These methods are one of the few to be simultaneously interpretable, non-linear, and easy-to-use. 

Random Forests have proven to be particularly effective%
. In a study of over one-hundred  datasets%
, Random Forests were found to be one of the best performing approaches --- even when no hyperparameter tuning is done \cite{JMLR:v15:delgado14a}. 
XGBoost, a variant of gradient boosting, has been used in the winning solutions to over half of recent Kaggle competitions \cite{xgboost}. 

Tree-based algorithms also provide a rare %
degree of interpretability. 
Single trees within an ensemble can be printed in a  human-readable form, allowing the immediate extraction of the decision process. %
Further still, there are numerous ways to extract feature importance scores from any tree-based approach \cite{NIPS2013_4928,breiman2003manual}. Being able to understand how a model reaches its decision is of special utility when we desire fair decision algorithms, as it gives us a method to double-check that the model appears to be making reasonable judgments. 
This interpretability has already been exploited in prior work to understand black-box models~\cite{Hall2017}.

Given the wide-ranging benefits and successes of tree-based learning, it is surprising that no prior work has focused on designing fair decision tree induction methods.
Other methods for constructing fair models will be reviewed in \autoref{sec:related_work}. 
In \autoref{sec:fair_forests} we propose, to the best of our knowledge, the first fair decision tree induction method. Our design is simple, requiring only minimal changes to existing tree induction code, thereby retaining the desirable property that the trees tend to "just work" %
without  hyperparameter tuning. Our experimental methodology is discussed in \autoref{sec:methodology}, including the introduction of novel fairness measures which are suitable for use with multinomial and continuous attributes. Finally, experimental results are summarized in \autoref{sec:experiments}, including a new experimental procedure to evaluate fair algorithms against all possible features rather than single protected attributed. We end with our conclusions in \autoref{sec:conclusion}. 

\section{Related Work} \label{sec:related_work}

One approach to building fair classifiers is based on data alteration, where the original corpus is altered to remove or mask information about the protected attribute. Some of the first work in fairness learning followed this approach, and attempted to make the minimum number of changes that removed the discriminative protective information \cite{Kamiran2009}. Others have attempted to re-label the data points to ensure a fair determination \cite{Luong:2011:KIS:2020408.2020488}. 

Another approach is to regularize the model in such a way that it is penalized for keeping information that allows it to discriminate against the protected feature. Some of the earliest work was to develop a fair version of Naive Bayes algorithm \cite{Calders:2010:TNB:1842547.1842562}. Others have taken to creating a differentiable regularization term, and applying it to models such as Logistic and Linear Regression \cite{Kamishima:2011:FLT:2117693.2119563,Bechavod2017,Berk2017,Calders2013}. Our new fair induction algorithm is a member of this group of regularization-based approaches, but unlike prior works has no parameters to tune.

One final group of related approaches is to build new representations, which mask the protected attribute \cite{Dwork:2012:FTA:2090236.2090255}. The use of neural networks have become popular for this task, such as variational auto encoders \cite{Louizos2016} and adversarial networks \cite{Edwards2016}. One of the seminal works in this field used an autoencoder with three separate terms in the loss \cite{pmlr-v28-zemel13}, and provides one of the largest comparisons on three now-standard datasets. We replicate their evaluation procedure in this work. 

There is an important commonality in all of these prior works. The research is done with respect to datasets and attributes where there is a prior normative expectation of fairness. These are problems usually of social importance, and protected attributes are intrinsic characteristics like age, gender and nationality. But what if focusing on such problems has inadvertently biased the development of fair research? The mechanism for inducing fairness should work for any attribute, not just those that align with current societal norms, and must not be over-fit to the protected attributes used in research. We evaluate our approach with respect to every possible feature choice, to ensure that the mechanism of producing fairness is not over-fit to the data.

\section{Fair Forests} \label{sec:fair_forests}

We propose a simple regularization approach to constructing a fair decision tree induction algorithm. This is done by altering the way we measure the information gain $G(T,a)$, where $T$ is a set of training examples, and $a$ is the attribute to split on. We will denote the set of points in each of the $k$ branchs of the tree as $T_{i \ldots k}$. This normally is combined with an impurity measure $I(T)$, to give us 
\begin{equation}
G(T,a) = I(T) - \sum_{\forall T_i \in \text{splits}(a)} \frac{\left|T_i\right|}{|T|} \cdot I(T_i)
\end{equation}

The information gain scores the quality of a splitting attribute $a$ by how much it reduces impurity compared to the current impurity. The larger the gain, the more pure the class labels have become, and thus, should improve classification performance. In the \textsc{cart} algorithm, the Gini impurity \eqref{eq:gini} is normally used for categorical targets. 
\begin{equation} \label{eq:gini}
I_{\text{Gini}}(T) = 1 - \sum_{\forall T_i \in \text{splits}(\text{label})} \left( \frac{\left|T_i\right|}{|T|}\right)^2
\end{equation}

This greedy approach to feature selection has proven effective for decades, helping to cement the place of tree-based algorithms as one of the most popular learning methods. However, this does not take into account any notion of fairness, which we desire to add. In this work we do so by altering the information gain scoring itself, leaving the whole of the tree induction process unaltered. 

We begin by noting we need to make two slight alterations for our approach. First, we will use the Impurity score to measure both the class label, and now additionally the protected attribute under consideration. We will denote these two cases as $I^l$, and $I^a$, and the Gain with respect to the label and protected attribute as $G^l$ and $G^a$ respectively. Additionally, we will impose the constraint that the impurity measure must return a value normalized to the range of $[0, 1]$. For the Gini measure this becomes 
\begin{equation} \label{eq:gini_new}
I^a_{\text{Gini}}(T) = \frac{1 - \sum_{\forall T_i \in \text{splits}(a)} \left( \frac{\left|T_i \right|}{|T|}\right)^2}{1-\left|\text{splits}(a)\right|^{-1} }
\end{equation}

We require that the impurity score $I^a(\cdot)$ produce a normalized score so that we can compare scores on a similar scale range, regardless of which features are selected. We then use this to define a new fair gain measure $G_{\text{fair}}(T, a)$, which seeks to balance predictive accuracy with the fairness goal with respect to some protected attribute $a_f$. 
\begin{equation} \label{eq:gain_fair}
G_{\text{fair}}(T, b) = G^l(T, b) - G^{a_f}(T,b)
\end{equation}
Intuitively,  \eqref{eq:gain_fair} will discourage the selection of any feature correlated with both the protected attribute and the target label. It remains possible for such a feature to still be selected if no other feature is better suited. 

\subsection{Gain for Numeric Features}

To our knowledge, no work has yet explored making a continuous feature the protected attribute. We can derive this naturally in our new fair induction framework. In \textsc{cart}, trees' numeric target variables are optimized by finding the binary split that minimizes the weighted variance between each split. We use this same notion to define a gain $G_{r}(T, a)$ that is used when either the predictor or protected attribute is continuous. 

Because we are interested in fairness, we look at changes in the mean value of the splits compared to their parent. Even if variances differ, if they retain similar means the impact on the fairness is minimal. To produce a scaled value, we look at the number of standard deviations from the previous mean is for each of the new splits, and assume that being more than three standard deviations is the maximum violation. 
This gain is defined  in \eqref{eq:gain_continuous}, where $\sigma_{b,T_i}$ indicates the standard deviation of attribute $b$ for all datums in the set $T_i$, and $\mu_{b, T_i}$ has the same meaning but for the mean of the subset.
\begin{equation}\label{eq:gain_continuous}
G_{r}(T, b) = 1 - \frac{1}{3}\sum_{T_i \in \text{split}} \frac{|T_i|}{|T|} \min\left(\frac{|\mu_{b,T}-\mu_{b,T_i}|}{\sigma_{b,T}} ,3\right)
\end{equation}
We emphasize that the standard deviation of the parent $T$ is used, not that of any sub-population $T_i$. This is because we want to measure drift with respect to the current status. Re-writing the continuous splitting criteria in this fashion also produces a score normalized to the range $[0, 1]$. We can now continue to use the $G_\text{fair}(T, b)$ function with continuous attributes as either the label target, or the protected attribute. 

This framework now gives us a means to induce decision trees, and thus build Random Forests, for all scenarios: classification and regression problems, and protected features either nominal or numeric. We emphasize that this approach to regularizing the information gain has no tunable parameters as given. This is to keep with the general utility of decision trees in that they often "just work."

While adjusting hyperparameters such as maximum tree depth may be used to improve classification accuracy, the results of a decision tree are often effective without any kind of parameter tunning. This is important for practical use and adoption. Many fairness based systems require an additional two to three hyperparameters to tune \cite{Kamishima:2011:FLT:2117693.2119563,Bechavod2017,pmlr-v28-zemel13}, on top of whatever hyperparameters come with the original model. This increases the computational requirements in practice, especially when used with a classic grid-search approach. 

\section{Methodology} \label{sec:methodology}

There is currently considerable discussion about what it means for a machine learning model to be fair, which metrics should be used, and whether or not they can be completely optimized \cite{Skirpan2017,Garcia-Martin2017,Hardt2016}. 

We choose to use the same evaluation procedure laid out by \citet{pmlr-v28-zemel13}. This makes our results comparable with a larger body of work, as their approach and metrics have been widely used through the literature \cite{Landeiro:2016:RTC:3015812.3015840,Bechavod2017,Dwork2017,Calders2013}. We present both of their metrics --- Discrimination and Inconsistency\footnote{\citeauthor{pmlr-v28-zemel13} refer to their metric as `consistency,' but define it in a way that only makes sense for classification. We use $\mathrm{Inconsistency} = 1-\mathrm{Consistency}$. This form is applicable to both classification and regression tasks.}
--- in a manner compatible with both classification and regression problems, while also extending Discrimination to a broader set of scenarios. We will also discuss the datasets used, their variants tested, and the models we will evaluate. 

\subsection{Metrics}

The first metric we will consider is the Discrimination of the model, measured by the average difference between the average predicted scores for each attribute value.%
\begin{equation} \label{eq:discrim}
\text{Discrimination} = \left| \frac{\sum_{x_i \in T_{a_p}} \hat{y}_i}{|T_{a_p}|} - \frac{\sum_{x_i \in T_{\neg a_p}} \hat{y}_i }{|  T_{\neg a_p} |} \right|
\end{equation}
Discrimination measures a macro-level quality of fairness, as such it is sometimes termed "group fairness." However, the definition in \eqref{eq:discrim} is limited to only binary protected attributes. For this work, we will also look at a generalization of Discrimination to $k$-way categorical variables. This is done by re-formulating Discrimination to consider the sub-population differences from the global mean. This is equivalent to the original definition when $k=2$, and is given by \eqref{eq:discrim_k}. (See the Appendix for a proof of equivalence.)
\begin{equation} \label{eq:discrim_k}
\text{Discrimination} = \frac{2}{k} \sum_{i = 1}^k \left| \frac{\sum_{x_j \in T} \hat{y}_j}{|T|} - \frac{\sum_{x_j \in T_i} \hat{y}_j}{|T_i|} \right|
\end{equation}

We will also consider the discrimination with respect to a continuous variable. With $a_p$ denoting a protected continuous attribute, let $x_i(a_p)$ be the value of feature $a_p$ for datum~$x_i$. We will then define our new Maximum Discrimination (MaxD) metric as the largest discrimination score achieved for some binary split of $a_p$ by some threshold $t$. This is given in equation \eqref{eq:discrim_r}, and gives us a concise definition extending Discrimination to regression tasks. When a continuous attribute is manually discretized into a binary problem, as is done in prior work, we obtain by definition that MaxD $\geq$ Discrimination. 
\begin{equation} \label{eq:discrim_r}
\text{MaxD} = \argmax_t \left| \frac{\sum_{x_i(a_p) < t} \hat{y}_i}{|x_i(a_p) < t|} - \frac{\sum_{x_i(a_p) \geq t} \hat{y}_i}{|x_i(a_p) \geq t|} \right|
\end{equation}

Given our novel extensions of the Discrimination scores \eqref{eq:discrim_k} and \eqref{eq:discrim_r}, we can evaluate this property for any feature. Importantly though, these metrics are population level measures of fairness. Satisfying the Discrimination metric does not guarantee that no bias exists. To measure the potential for bias within sub-populations of the data set, we look at the Inconsistency metric \eqref{eq:inconsistency}. 
\begin{equation} \label{eq:inconsistency}
\text{Inconsistency} = \frac{1}{N} \sum_{i = 1}^N \left| \hat{y}_i - \frac{1}{k} \sum_{j \in k\text{-NN}(x_i)} \hat{y}_j\right|
\end{equation}
Inconsistency compares the prediction of the model with that of nearby points, and is sometimes referred to as "individual fairness." This is under the assumption that nearby points should produce similar predictions, and is optimized when the score is as close to zero as possible.%

Discrimination and Inconsistency are both evaluating the fairness of a model, and hence do not consider the true supervised label $y$. Maximizing fairness involves minimizing these two scores, at a potential cost to the model's predictive utility. We measure the predictive utility of each model with accuracy or Root Mean Squared Error (RMSE) for classification and regression problems respectively. For classification problems, we also consider the \textit{Delta} metric, where $\text{Delta} =\text{Accuracy} - \text{Discrimination}$. 

For corpora with a test set, these metrics will all be evaluated on the given test set. Otherwise, we will evaluate these scores on 10-fold cross validation. For Inconsistency, we will measure it using nearest neighbors from all folds, but using the predicted scores obtained from cross validation. This is in keeping with prior work \cite{pmlr-v28-zemel13}.

\subsection{Data Sets}

To evaluate our work, we will use three classification datasets used by \citet{pmlr-v28-zemel13}, the German Credit, Adult, and Heritage Health datasets. For regression we will also use the Health dataset, which was originally a regression problem (how many days will someone stay in the hospital?) that was converted to classification (will they stay one or more days?).

\autoref{tbl:dataset_summary} summarizes the size, protected attribute, feature count, and task type for each dataset. For the German  and Health datasets, the protected attribute \textit{age} is originally encoded as a numeric feature, but, because prior work did not support continuous protected attributes, they converted it to a binary categorical feature. We replicate this in our work, but will also investigate using the original continuous version of \textit{age}. 
\begin{table}[!htbp]
\centering
\caption{Summary of the datasets used.}
\label{tbl:dataset_summary}
\begin{adjustbox}{max width=\columnwidth}
\begin{tabular}{@{}lrcrc@{}}
\toprule
Dataset         & \multicolumn{1}{c}{Samples} & Protected            & \multicolumn{1}{c}{Featuers} & Task \\ \midrule
German Credit   & 1,000                       & Age $\geq$ 25        & 20                           & Good/Bad Credit    \\
Adult Income    & 45,222                      & Male/Female          & 14                           & Income $\geq$ 50k    \\
Heritage Health & 147,471                     & Age $\geq$ 65        & 149                          & Stay $\geq$ 1 day   \\
Heritage HealthR & 147,471                     & Age $\geq$ 65        & 149                          & Days in stay    \\
\bottomrule
\end{tabular}
\end{adjustbox}
\end{table}

\begin{table*}[!t]
\centering
\caption{For each classification task, we show Accuracy, Delta, Discrimination, and Inconsistency, in that order. Scores are for our new method and prior work. Best results shown in \textbf{bold}, second best in \textit{italics}. }
\label{tbl:fair_results_c}
\begin{adjustbox}{max width=\textwidth}
\begin{tabular}{@{}lcccccccccccc@{}}
\toprule
      & \multicolumn{4}{c}{German}                                            & \multicolumn{4}{c}{Adult}                                             & \multicolumn{4}{c}{Health}                                            \\ 
      \cmidrule(l){2-5} \cmidrule(l){6-9}  \cmidrule(l){10-13} 
      & Acc             & Delta           & Discrim         & Incon           & Acc             & Delta           & Discrim         & Incon           & Acc             & Delta           & Discrim         & Incon           \\ \midrule
DT    & 0.6890          & 0.6509          & 0.0381          & 0.2140          & 0.8364          & 0.4801          & 0.3563          & 0.4417          & 0.8404          & 0.8196          & 0.0207          & 0.2062          \\
\dtf  & \textit{0.6990} & 0.6908          & 0.0082          & \textit{0.0070} & 0.7511          & 0.7444          & 0.0067          & \textit{0.0033} & \textbf{0.8474} & \textit{0.8473} & \textit{0.0001} & \textit{0.0001} \\
RF    & 0.6970          & \textit{0.6911} & 0.0059          & 0.0020          & \textbf{0.8501} & 0.5463          & 0.3038          & 0.3944          & 0.8472          & 0.8464          & 0.0007          & 0.0005          \\
\rff  & \textbf{0.7000} & \textbf{0.7000} & \textbf{0.0}    & \textbf{0.0}    & 0.7530          & \textit{0.7530} & \textbf{0.0}    & \textbf{0.0}    & \textbf{0.8474} & \textbf{0.8474} & \textbf{0.0}    & \textbf{0.0}    \\ \cline{2-13} 
\nbf  & 0.6888          & 0.6314          & 0.0574          & 0.3132          & \textit{0.7847} & \textbf{0.7711} & 0.0136          & 0.4366          & 0.6878          & 0.5678          & 0.1200          & 0.4107          \\
LR    & 0.6790          & 0.5517          & 0.1273          & 0.3050          & 0.6787          & 0.4895          & 0.1892          & 0.2703          & 0.7547          & 0.6482          & 0.1064          & 0.2767          \\
\lrf  & 0.5953          & 0.5842          & 0.0111          & 0.1284          & 0.6758          & 0.6494          & 0.0264          & 0.2234          & 0.7212          & 0.7038          & 0.0174          & 0.3777          \\
LFR   & 0.5909          & 0.5867          & 0.0042          & 0.0592          & 0.7023          & 0.7018          & \textit{0.0006} & 0.1892          & 0.7365          & 0.7365          & \textbf{0.0000} & \textbf{0.0000} \\ \bottomrule
\end{tabular}
\end{adjustbox}
\end{table*}

\subsection{Models Evaluated}

When listing results, we will compare with standard \textsc{cart} decision trees (DT) and Random Forests (RF). Our fair variants of these methods will be denoted as \dtf and \rff. 

Since our new fair tree induction can directly protect the original non-discretized form, we also evaluate in that manner. Models $\text{DT}^F_c$ and $\text{RF}^F_c$ indicate a fair decision tree and Random Forest trained to protect the continuous age attribute. When we do this, we will continue to evaluate the models' Discrimination with the originally proposed threshold. 

From \citet{pmlr-v28-zemel13}, we compare against their proposed Learning Fair Representations (LFR) approach and their baseline approaches: Logistic Regression, fair Logistic Regression (\lrf) \cite{Kamishima:2011:FLT:2117693.2119563} and  fair Naive Bayes (\nbf) \cite{Kamiran2009}.

\section{Experiments} \label{sec:experiments}

In this section we present the results of our experiments. We remind the reader that for all experiments, we perform no parameter turning for any of our tree-based models. This is in line with practical use, and is a benefit for users in both runtime and simplicity. In these experiments we will show our Fair Forests can be used in the standard classification scenario with a binary protected attribute. In addition, we can use a continuous protected attribute and achieve similar results, and apply both methods to a numeric prediction target. All code was written in Java using the JSAT library~\cite{JMLR:v18:16-131}. 

\subsection{Binary Target, Binary Protected}

For the classification tasks, the results for our various decision tree variants can be seen compared to the  baselines in \autoref{tbl:fair_results_c}. We can see that our new Fair Forests win in almost every metric.

Looking at just the tree-based results, we can make two interesting observations. First, that the ensembling and random feature sub-sampling used by Random Forests appears to improve the fairness of \textsc{cart} trees, both when they do and do not consider our fairness regularization. This is a positive indication for the general use of Random Forests compared to single decision trees. Second, that our fairness regularizer can actually \textit{improve} accuracy. This was observed on the German and Health datasets. We do not observe this phenomena with any other fairness approach. While a positive result, we caution that this should not be a general expectation. It is always possible that the protected attribute may truly be predictive of the target task. In such cases we would expect performance to decrease, which we observe on the Adult income dataset. 

This result is also important when we contrast to the Logistic Regression model and its fair variant. On every dataset tested, the fair variant of LR has worse predictive accuracy than the standard model. Our fair trees do not suffer in the same way, indicating they are a more robust approach to building fair models. 

The baseline results shown from \citet{pmlr-v28-zemel13} required a grid-search, and were selected to maximize the Delta score. In this regard our Fair Forests almost dominate the table. The Fair Forest is second best only once to \nbf on the Adult Income corpus, with a relative difference of merely 2.3\%. \nbf achieves this by obtaining higher accuracies, but also a higher Discrimination. 

On both measures of fairness, Discrimination and Inconsistency, our fair Random Forest dominates the table with empirical zeros. The best non-tree approach in this regard is the LFR algorithm, which obtains empirical zeros on the Health dataset and near-zeros for Discrimination on the German and Adult datasets. However, LFR's Inconsistency increases to 0.06 and 0.19 for each respectively.

\subsection{Binary Target, Continuous Protected}

In all prior literature we are aware of, the protected attribute is always presented as a binary feature. Our fair tree induction approach allows us to mark a continuous feature as protected directly, without first having to discretize it. We can test this ability with the German and Health datasets, where the protected attribute (\textit{age}) is originally a numeric feature. The results comparing this approach with the classic binary \textit{age} attribute are shown in \autoref{tbl:results_c_regProt}. In this table  Discrimination is based on the original thresholds used to binarize the \textit{age}. 

\begin{table}[!htbp]
\centering
\caption{Classification results on German and Health datasets, where the protected \textit{age} attribute is left as a numeric feature, rather than being converted to a binary categorical one. Best results in \textbf{bold}, second best in \textit{italics}. }
\label{tbl:results_c_regProt}
\begin{adjustbox}{max width=\columnwidth}
\begin{tabular}{@{}lcccccccc@{}}
\toprule
      & \multicolumn{4}{c}{German}                                            & \multicolumn{4}{c}{Health}                                            \\ 
      \cmidrule(l){2-5} \cmidrule(l){6-9}
      & Acc             & MaxD           & Discrim         & Incon           & Acc             & MaxD           & Discrim         & Incon           \\ \midrule
\dtf  & \textit{0.6990} & 0.0216          & 0.0082          & 0.0070          & \textbf{0.8474} & \textit{0.0001} & \textit{0.0001} & \textit{0.0001} \\
\dtfc & 0.6960          & \textit{0.0054} & \textit{0.0047} & \textit{0.0040} & \textbf{0.8474} & \textbf{0.0}    & \textbf{0.0}    & \textbf{0.0}    \\
\rff  & \textbf{0.7000} & \textbf{0.0}    & \textbf{0.0}    & \textbf{0.0}    & \textbf{0.8474} & \textbf{0.0}    & \textbf{0.0}    & \textbf{0.0}    \\
\rffc & \textbf{0.7000} & \textbf{0.0}    & \textbf{0.0}    & \textbf{0.0}    & \textbf{0.8474} & \textbf{0.0}    & \textbf{0.0}    & \textbf{0.0}    \\ \bottomrule
\end{tabular}
\end{adjustbox}
\end{table}

Here we can see that the Random Forest using the continuous \textit{age} (\rffc) and the one using binary \textit{age} (\rff) have equivalent performance. This appears to be a net effect of the added fairness Random Forests naturally provide. In this case it becomes more informative to look at the results from the standard decision tree, where non-zero Discrimination still occurs. 

For both \dtf and \dtfc, we can see they continue to  reduce the Discrimination and Inconsistency with respect to the original decision tree approach. In these cases, \dtfc appears to uniformly outperform \dtf in regards to the fairness metrics, with only a 0.003 change in accuracy. On the German dataset we can also see that \dtfc has improved upon the MaxD score from the \dtf, dropping from 0.0216 down to 0.0054. This is reasonable to expect, as \dtf is optimizing fairness with respect to a specific value of \textit{age}, where \dtfc is attempting to be fair with respect to all \textit{age} values. 

Further reading of the table indicates the MaxD score for \dtfc (0.005) is smaller than the Discrimination score of the \dtf approach (0.008). This means \dtfc has a greater degree of fairness with respect to age for all possible age splits, than \dtf does with regard to its single age split of interest. We explain this result by noting that \dtf's single split focus at \textit{age} $\geq 25$ means discrimination can occur in nearby age ranges (e.g., 26-30, or 21-24), and this permissible "border" discrimination can generalize into the test set. Ultimately, while protecting the binary \textit{age} attribute works well compared to the naive DT in \autoref{tbl:fair_results_c}, these results demonstrate the benefit of protecting the original numeric attribute: we can provide better fairness with respect to the threshold of interest, as well as every possible other threshold.

\subsection{Continuous Target}

One of the benefits of Decision Trees or Random Forests is that they can be applied to both classification and regression problems. In this section we will show that our fair induction strategy improves fairness in such scenarios, both when protecting on a continuous or a binary attribute. We do this using the original version of the Health dataset (see \autoref{tbl:results_healthr}).
\begin{table}[!htbp]
\centering
\caption{Results using the Regression version of the Health dataset, where Mean Squared Error is the target metric. Best results shown in \textbf{bold}, second best in \textit{italics}. }
\label{tbl:results_healthr}
\begin{tabular}{@{}lllll@{}}
\toprule
     & MSE            & Discrim         & MaxD            & Incon           \\ \midrule
LR   & 2.722          & 0.0011          & 0.0024          & 0.4163          \\
DT   & 2.904          & 0.0006          & 0.0057          & 1.0285          \\
\dtf  & \textbf{2.662} & \textit{0.0005} & \textbf{0.0005} & 0.0964          \\
\dtfc & \textit{2.663} & \textbf{0.0003} & \textit{0.0007} & \textbf{0.0695} \\
RF   & 2.735          & 0.0021          & 0.0022          & 1.0464          \\
\rff  & 2.664          & 0.0006          & 0.0006          & 0.1366          \\
\rffc & 2.664          & 0.0007          & 0.0010          & \textit{0.0690} \\ \bottomrule
\end{tabular}
\end{table}

When phrased as a regression problem, we see lower Discrimination scores for both the standard Decision Tree and the Random Forest, which leaves little room for improvement. When comparing Discrimination against the binary \textit{age} threshold (\textit{age} $\geq$ 65), and the Maximum Discrimination against \textit{age}, we see the fair variants of our algorithms perform better than their non-fair counterparts. While the \dtf and \dtfc happen to perform slightly better than their counterparts \rff and \rffc, the differences are in an epsilon range.  Either way, these results show that we can use our approach for regression problems and protect both categorical and continuous attributes. 

\subsection{Visualizing the Impact of Fairness}

\begin{figure} [htbp]%
    \centering
    \begin{adjustbox}{max width=\columnwidth}
    \begin{tikzpicture}[]
        \begin{axis}[
        	height=0.4\textheight,
            width=\columnwidth,
            xbar,
            xmax=1.0,
            bar width=3.0,
            y tick label style={font=\small,rotate=45, anchor=south east, inner sep=0mm},
            ytick=data,
            grid=major,
            symbolic y coords={duration,credit-amount,installment-commitment,residence-since,age,existing-credits,num-dependents,checking-status,credit-history,purpose,savings-status,employment,personal-status,other-parties,property-magnitude,other-payment-plans,housing,job,own-telephone,foreign-worker},
            xlabel={Relative Importance},
            enlarge x limits={abs=0},
			enlarge y limits=0.03,
            reverse legend,
        ]
        \addplot table [x=raw, y=feature,col sep=comma] {german_import.csv};
        \addplot table [x=fair, y=feature,col sep=comma] {german_import.csv};
        \legend{Standard,Fair}
        \end{axis}
    \end{tikzpicture}
    \end{adjustbox}
    \caption{Feature importance from German dataset. }
\label{fig:german_importance}
\end{figure}
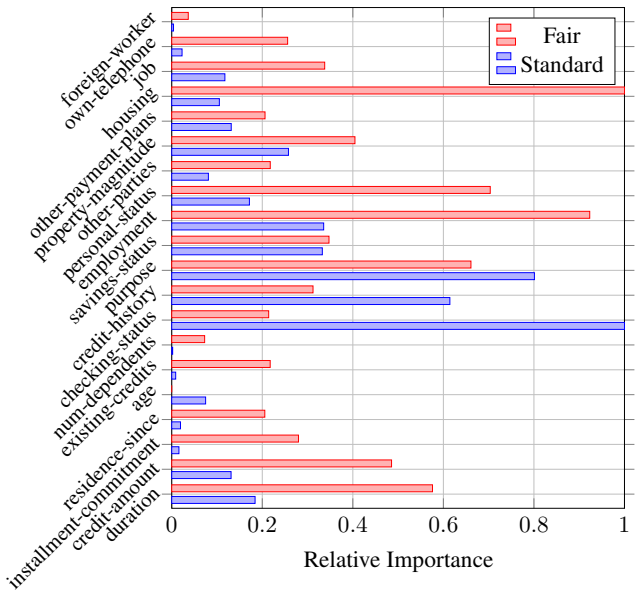

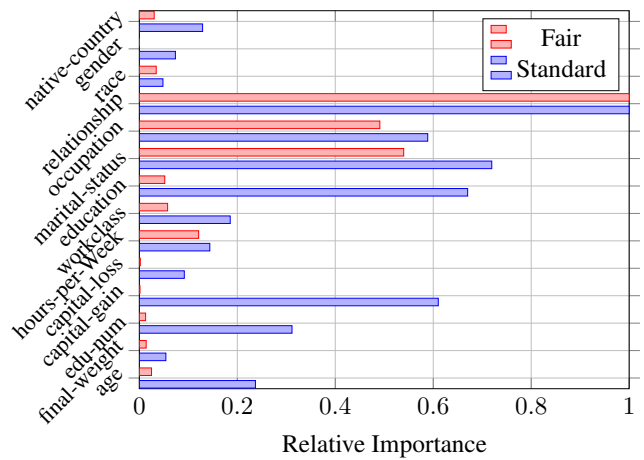
\begin{figure} [tbp]%
    \centering
    \begin{adjustbox}{max width=\columnwidth}
    \begin{tikzpicture}[]   
        \begin{axis}[
        	height=0.30\textheight,
            width=\columnwidth,
            xbar,                                 
            xmax=1.0,
            bar width=3.0,
            y tick label style={font=\small,rotate=45, anchor=south east, inner sep=0mm},
            ytick=data,
            grid=major,
			symbolic y coords={age,final-weight,edu-num,capital-gain,capital-loss,hours-per-Week,workclass,education,marital-status,occupation,relationship,race,gender,native-country},
            xlabel={Relative Importance},
            enlarge x limits={abs=0},
			enlarge y limits=0.03,
            reverse legend,
        ]
        \addplot table [x=raw, y=feature,col sep=comma] {adult_income_import.csv};
        \addplot table [x=fair, y=feature,col sep=comma] {adult_income_import.csv};
        \legend{Standard,Fair}
        \end{axis}
    \end{tikzpicture}
    \end{adjustbox}
    \caption{Feature importance from Adult dataset. }
\label{fig:adult_importance}
\end{figure}

One of the benefits of tree-based approaches to prediction is the ability to interpret the models. In particular, we note that one can measure the relative importance of a feature using a variety of approaches. Using the Mean Decrease in Impurity (MDI) measure \cite{NIPS2013_4928}, we show the relative importance of features on the German and Adult datasets. These are shown in \autoref{fig:german_importance} and \autoref{fig:adult_importance} respectively, where "Fair" is the relative importance of features used by our Fair Forest induction algorithm and "Standard" indicates the normal Random Forest induction process using \textsc{cart}-style trees. These results allow us to see that our simple regularizer can have a wide range of impact, depending on the dataset and protected attribute. 

On the German dataset in \autoref{fig:german_importance}, we see a dramatic change in what the model considers importance, with the the most important variable being \textit{checking-status}  under the Standard model but \textit{housing} under the Fair model. For almost all features in this corpus, we see a reversing of importance: if it was important under the naive model, it becomes less-so under the Fair model, and vice versa. The only exception to this being the \textit{savings-status} attribute, and to some degree, \textit{property-magnitude}. 

The Adult dataset has a markedly different and surprising behavior. Under both the Fair and Naive model, the \textit{relationship} attribute continues to be the most important. However, the Fair model dramatically reduces the relative importance of most other features.
Many of these (e.g. \textit{capital-loss}, \textit{capital-gain}, \textit{education}) would likely be features we expect to reliably predict the target attribute, Income.
While our intuition may be that these variables should be unbiased and naturally fair predictors, the underlying distribution of \textit{this dataset} indicates they were too highly correlated with the protected Gender attribute, and thus were rarely selected for use. 

We expect that the ability to perform such investigation into feature importance pre/post fairness will become a valuable tool for those who wish to build fair models in production environments. Changes in feature importance can give us underlying insights into non-linear correlations that would escape simple analysis. The information itself may allow a decision maker to discover deficiencies or unintended biases in their data collection process, based on these unexpected changes. For example, the non-use of the \textit{capital-gain}/\textit{loss} features may tell us that we need to collect more data specifically from women with capital investments. 

\subsection{Fairness vs the Mechanism}

We now evaluate the ability of our model to reduce Discrimination for every attribute individually, across each dataset. This helps us to determine that our approach is not overly specific to the choice of attributes such as age and gender. To our knowledge this is the first such evaluation in the fairness literature. 

First we train a standard Random Forest, and measure the Discrimination for each attribute using \eqref{eq:discrim_k} or \eqref{eq:discrim_r} as appropriate. From these we record the average and standard deviation of the "Raw" discrimination. Then we train a new Fair Forest $D$ times for $D$ features,  testing the model when each feature is selected as the protected attribute. We then measure the Discrimination of the protected feature  and the accuracy of the resulting model. The mean and standard deviation are then calculated from the protected feature Discriminations.  The results of this are shown in \autoref{tbl:discrim_all}.

\begin{table}[tbp]
\centering
\caption{Discrimination statistics for all features in each dataset. First row is the Discrimination without any protection. The second row shows Discrimination when protecting each feature individually, and third row shows the associated model accuracy.}
\label{tbl:discrim_all}
\begin{adjustbox}{max width=\columnwidth}
\begin{tabular}{@{}lcccccc@{}}
\toprule
              & \multicolumn{2}{c}{German} & \multicolumn{2}{c}{Adult} & \multicolumn{2}{c}{Health} \\ 
              \cmidrule(l){2-3} \cmidrule(l){4-5} \cmidrule(l){6-7}
              & $\mu$      & $\sigma$      & $\mu$      & $\sigma$     & $\mu$     & $\sigma$       \\ \midrule
Raw Discrim   & 0.0081     & 0.0137        & 0.2971     & 0.1652       & 0.0066    & 0.0101         \\
Prot. Discrim  & 0.0000     & 0.0000        & 0.1253     & 0.0776       & 0.0000    & 0.0000         \\
Prot. Accuracy & 0.7000     & 0.0000        & 0.8044     & 0.0108       & 0.8474    & 0.0000          \\ \bottomrule
\end{tabular}
\end{adjustbox}
\end{table}

Across all three datasets and every feature, the Fair Forest approach was always able to decrease the Discrimination with respect to the protected attribute. For the German and Health datasets, it is able to reduce the Discrimination to zero for all features, and always results in the same accuracy. For the Adult dataset, the original protected attribute of Gender was the only attribute which could be reduced to a Discrimination of zero. The Adult dataset is the only one producing a wide impact in the amount of Discrimination removed, and the resulting accuracy of the model (decreasing from 0.85  down to 0.80 on average). 

\section{Conclusion} \label{sec:conclusion}

We have developed, to the best of our knowledge, the first fair variant of the Random Forest algorithm. This Fair Forest can be used for classification and regression problems, and protected $k$-category features as well as numeric attributes, a first in the fairness literature. In doing so we have extended the measure of discrimination to these cases. We have shown our method produces state-of-the art results on three common benchmark datasets, while requiring no parameter tuning to use, and is able to uniformly reduce Discrimination across any feature in each corpus. 

{\small
\bibliographystyle{aaai}
\bibliography{Mendeley}

\begin{thebibliography}{}

\bibitem[\protect\citeauthoryear{Bechavod and Ligett}{2017}]{Bechavod2017}
Bechavod, Y., and Ligett, K.
\newblock 2017.
\newblock {Learning Fair Classifiers: A Regularization-Inspired Approach}.
\newblock In {\em FAT ML Workshop}.

\bibitem[\protect\citeauthoryear{Berk \bgroup et al\mbox.\egroup
  }{2017}]{Berk2017}
Berk, R.; Heidari, H.; Jabbari, S.; Joseph, M.; Kearns, M.; Morgenstern, J.;
  Neel, S.; and Roth, A.
\newblock 2017.
\newblock {A Convex Framework for Fair Regression}.
\newblock In {\em FAT ML Workshop}.

\bibitem[\protect\citeauthoryear{Breiman \bgroup et al\mbox.\egroup
  }{1984}]{Breiman1984}
Breiman, L.; Friedman, J.; Stone, C.~J.; and Olshen, R.
\newblock 1984.
\newblock {\em {Classification and Regression Trees}}.
\newblock CRC press.

\bibitem[\protect\citeauthoryear{Breiman}{2001}]{Breiman2001}
Breiman, L.
\newblock 2001.
\newblock {Random forests}.
\newblock {\em Machine learning} 45(1):5--32.

\bibitem[\protect\citeauthoryear{Breiman}{2003}]{breiman2003manual}
Breiman, L.
\newblock 2003.
\newblock {Manual on setting up, using, and understanding random forests v4.0}.
\newblock {\em Statistics Department University of California Berkeley, CA,
  USA}.

\bibitem[\protect\citeauthoryear{Calders and
  Verwer}{2010}]{Calders:2010:TNB:1842547.1842562}
Calders, T., and Verwer, S.
\newblock 2010.
\newblock {Three Naive Bayes Approaches for Discrimination-free
  Classification}.
\newblock {\em Data Min. Knowl. Discov.} 21(2):277--292.

\bibitem[\protect\citeauthoryear{Calders \bgroup et al\mbox.\egroup
  }{2013}]{Calders2013}
Calders, T.; Karim, A.; Kamiran, F.; Ali, W.; and Zhang, X.
\newblock 2013.
\newblock {Controlling Attribute Effect in Linear Regression}.
\newblock In {\em 2013 IEEE 13th International Conference on Data Mining},
  71--80.
\newblock IEEE.

\bibitem[\protect\citeauthoryear{Chen and Guestrin}{2016}]{xgboost}
Chen, T., and Guestrin, C.
\newblock 2016.
\newblock {XGBoost: Reliable Large-scale Tree Boosting System}.
\newblock In {\em Proceedings of the 22nd ACM SIGKDD International Conference
  on Knowledge Discovery and Data Mining}.

\bibitem[\protect\citeauthoryear{Dwork \bgroup et al\mbox.\egroup
  }{2012}]{Dwork:2012:FTA:2090236.2090255}
Dwork, C.; Hardt, M.; Pitassi, T.; Reingold, O.; and Zemel, R.
\newblock 2012.
\newblock {Fairness Through Awareness}.
\newblock In {\em Proceedings of the 3rd Innovations in Theoretical Computer
  Science Conference}, ITCS '12,  214--226.
\newblock New York, NY, USA: ACM.

\bibitem[\protect\citeauthoryear{Dwork \bgroup et al\mbox.\egroup
  }{2017}]{Dwork2017}
Dwork, C.; Immorlica, N.; Kalai, A.~T.; and Leiserson, M.
\newblock 2017.
\newblock {Decoupled classifiers for fair and efficient machine learning}.
\newblock In {\em FAT ML Workshop}.

\bibitem[\protect\citeauthoryear{Edwards and Storkey}{2016}]{Edwards2016}
Edwards, H., and Storkey, A.
\newblock 2016.
\newblock {Censoring Representations with an Adversary}.
\newblock In {\em International Conference on Learning Representations (ICLR)}.

\bibitem[\protect\citeauthoryear{Fern{\'{a}}ndez-Delgado \bgroup et
  al\mbox.\egroup }{2014}]{JMLR:v15:delgado14a}
Fern{\'{a}}ndez-Delgado, M.; Cernadas, E.; Barro, S.; and Amorim, D.
\newblock 2014.
\newblock {Do we Need Hundreds of Classifiers to Solve Real World
  Classification Problems?}
\newblock {\em Journal of Machine Learning Research} 15:3133--3181.

\bibitem[\protect\citeauthoryear{Friedman}{2002}]{Friedman2002}
Friedman, J.~H.
\newblock 2002.
\newblock {Stochastic gradient boosting}.
\newblock {\em Computational Statistics {\&} Data Analysis} 38(4):367–378.

\bibitem[\protect\citeauthoryear{Garc{\'{i}}a-Mart{\'{i}}n and
  Lavesson}{2017}]{Garcia-Martin2017}
Garc{\'{i}}a-Mart{\'{i}}n, E., and Lavesson, N.
\newblock 2017.
\newblock {Is it ethical to avoid error analysis?}
\newblock In {\em FAT ML Workshop}.

\bibitem[\protect\citeauthoryear{Hall and Gill}{2017}]{Hall2017}
Hall, P., and Gill, N.
\newblock 2017.
\newblock {Debugging the Black-Box COMPAS Risk Assessment Instrument to
  Diagnose and Remediate Bias}.

\bibitem[\protect\citeauthoryear{Hardt, Price, and Srebro}{2016}]{Hardt2016}
Hardt, M.; Price, E.; and Srebro, N.
\newblock 2016.
\newblock {Equality of Opportunity in Supervised Learning}.
\newblock In {\em Advances in Neural Information Processing Systems 29 (NIPS
  2016)}.

\bibitem[\protect\citeauthoryear{Kamiran and Calders}{2009}]{Kamiran2009}
Kamiran, F., and Calders, T.
\newblock 2009.
\newblock {Classifying without discriminating}.
\newblock In {\em 2009 2nd International Conference on Computer, Control and
  Communication},  1--6.
\newblock IEEE.

\bibitem[\protect\citeauthoryear{Kamishima, Akaho, and
  Sakuma}{2011}]{Kamishima:2011:FLT:2117693.2119563}
Kamishima, T.; Akaho, S.; and Sakuma, J.
\newblock 2011.
\newblock {Fairness-aware Learning Through Regularization Approach}.
\newblock In {\em Proceedings of the 2011 IEEE 11th International Conference on
  Data Mining Workshops}, ICDMW '11,  643--650.
\newblock Washington, DC, USA: IEEE Computer Society.

\bibitem[\protect\citeauthoryear{Landeiro and
  Culotta}{2016}]{Landeiro:2016:RTC:3015812.3015840}
Landeiro, V., and Culotta, A.
\newblock 2016.
\newblock {Robust Text Classification in the Presence of Confounding Bias}.
\newblock In {\em Proceedings of the Thirtieth AAAI Conference on Artificial
  Intelligence}, AAAI'16,  186--193.
\newblock AAAI Press.

\bibitem[\protect\citeauthoryear{Louizos \bgroup et al\mbox.\egroup
  }{2016}]{Louizos2016}
Louizos, C.; Swersky, K.; Li, Y.; Welling, M.; and Zemel, R.
\newblock 2016.
\newblock {The Variational Fair Autoencoder}.
\newblock In {\em International Conference on Learning Representations (ICLR)}.

\bibitem[\protect\citeauthoryear{Louppe \bgroup et al\mbox.\egroup
  }{2013}]{NIPS2013_4928}
Louppe, G.; Wehenkel, L.; Sutera, A.; and Geurts, P.
\newblock 2013.
\newblock {Understanding variable importances in forests of randomized trees}.
\newblock In Burges, C.; Bottou, L.; Welling, M.; Ghahramani, Z.; and
  Weinberger, K., eds., {\em Advances in Neural Information Processing Systems
  26}.
\newblock  431--439.

\bibitem[\protect\citeauthoryear{Luong, Ruggieri, and
  Turini}{2011}]{Luong:2011:KIS:2020408.2020488}
Luong, B.~T.; Ruggieri, S.; and Turini, F.
\newblock 2011.
\newblock {k-NN As an Implementation of Situation Testing for Discrimination
  Discovery and Prevention}.
\newblock In {\em Proceedings of the 17th ACM SIGKDD International Conference
  on Knowledge Discovery and Data Mining}, KDD '11,  502--510.
\newblock New York, NY, USA: ACM.

\bibitem[\protect\citeauthoryear{Pedreshi, Ruggieri, and
  Turini}{2008}]{Pedreshi:2008:DDM:1401890.1401959}
Pedreshi, D.; Ruggieri, S.; and Turini, F.
\newblock 2008.
\newblock {Discrimination-aware Data Mining}.
\newblock In {\em Proceedings of the 14th ACM SIGKDD International Conference
  on Knowledge Discovery and Data Mining}, KDD '08,  560--568.
\newblock New York, NY, USA: ACM.

\bibitem[\protect\citeauthoryear{Quinlan}{1993}]{Quinlan1993}
Quinlan, J.~R.
\newblock 1993.
\newblock {\em {C4.5: Programs for Machine Learning}}, volume~1 of {\em Morgan
  Kaufmann series in {\{}M{\}}achine {\{}L{\}}earning}.
\newblock Morgan Kaufmann.

\bibitem[\protect\citeauthoryear{Raff}{2017}]{JMLR:v18:16-131}
Raff, E.
\newblock 2017.
\newblock {JSAT: Java Statistical Analysis Tool, a Library for Machine
  Learning}.
\newblock {\em Journal of Machine Learning Research} 18(23):1--5.

\bibitem[\protect\citeauthoryear{Skirpan and Gorelick}{2017}]{Skirpan2017}
Skirpan, M., and Gorelick, M.
\newblock 2017.
\newblock {The Authority of "Fair" in Machine Learning}.
\newblock In {\em FAT ML Workshop}.

\bibitem[\protect\citeauthoryear{Zemel \bgroup et al\mbox.\egroup
  }{2013}]{pmlr-v28-zemel13}
Zemel, R.; Wu, Y.; Swersky, K.; Pitassi, T.; and Dwork, C.
\newblock 2013.
\newblock {Learning Fair Representations}.
\newblock In Dasgupta, S., and McAllester, D., eds., {\em Proceedings of the
  30th International Conference on Machine Learning}, volume~28 of {\em
  Proceedings of Machine Learning Research},  325--333.
\newblock Atlanta, Georgia, USA: PMLR.

\end{thebibliography}
}

\appendix
\setcounter{section}{-1}
\section{Appendix}
\renewcommand{\thesection}{A}

\subsection{Proof of $k$-way Discrimination}

Here we prove that \eqref{eq:discrim} and \eqref{eq:discrim_k} are equivalent when $k=2$. 

To simplify, let the predictive mean for the points $T$ be $\mu$, and for each subset $T_i$ be $\mu_i$. Writing out for $k=2$ for \eqref{eq:discrim_k}, we get 
$$
\text{Discrimination} = \left| \mu - \mu_1 \right| + \left| \mu - \mu_2 \right|
$$

Setting \eqref{eq:discrim} and \eqref{eq:discrim_k} equal to each other we get
$$
 \left| \mu_1 - \mu_2 \right| =  \left| \mu - \mu_1 \right| + \left| \mu - \mu_2 \right|
$$

Assume, without loss of generality, that $\mu_1 > \mu > \mu_2$. In this case we can use the absolute value to re-write as 
$$
 \left| \mu_1 - \mu_2 \right| =  \left| \mu_1 - \mu \right| + \left| \mu - \mu_2 \right|
$$
and then equivalently simplify as
$$
 \mu_1 - \mu_2 =   (\mu_1 - \mu) +  (\mu - \mu_2) 
$$
$$
 \mu_1 - \mu_2 =   \mu_1   - \mu_2 +  \mu - \mu 
$$
$$
 \mu_1 - \mu_2 =   \mu_1   - \mu_2
$$. 

Thus we obtain the same solution given a fixed ordering of the $\mu$s. The absolute value operation allows us to re-order the contained terms to match any distinct order of $\mu$s. For the case that $\mu_1 = \mu_2$, then it must be that $\mu = \mu_1$, and all terms will zero out. Therefor, we prove that \eqref{eq:discrim} = \eqref{eq:discrim_k}. 

\subsection{Why Define Gain on Mean over Variance?}

We take a moment to further expound upon why we have re-written the gain of numeric attributes with respect to the difference in means, when the original \textsc{cart} approach uses a criteria based on a reduction in variance. This original \textsc{cart} splitting condition can be defined by 

\begin{equation} \label{eq:cart_r}
\argmin_{t} \frac{1}{t} \sum_{i=1}^t \sum_{j=1}^t \frac{1}{2}(x_i - x_j)^2 + \frac{1}{n\!-\!t} \sum_{i=t+1}^n \sum_{j=t+1}^n \frac{1}{2}(x_i - x_j)^2
\end{equation}

Where $t$ is the splitting point, and $x_i \leq x_{i+1}$. This measures the sum of weighted variances between the two sets of points. This could be converted into our normalized gain form. Using the same notation, this would be 

\begin{equation}\label{eq:gain_r_variance}
G_{v}(T, b) = 1 - \sum_{T_i \in \text{split}} \frac{|T_i|}{|T|} \frac{\sigma_{b,T_i}^2}{\sigma_{b,T}^2}
\end{equation}

Using \eqref{eq:gain_r_variance} would have the desirable property of being equivalent to the solution found by \eqref{eq:cart_r}, and thus producing the same trees when there is no protected attribute. The problem with using this approach is that it does not align with our fairness goal, and can fail to produce fair trees when protecting numeric attributes. 

\begin{figure}[!htb]
\begin{tikzpicture}
\begin{axis}
	[
    	ylabel = Splitting Attribute,
        xlabel = Protected Attribute Value,
    	legend pos=north east
    ]
    \addplot[
		scatter/classes={
			a={mark=square*,blue},%
			b={mark=triangle*,red}%
		},
		scatter,only marks,
		scatter src=explicit symbolic]
	table[x=x,y=y,meta=label]{gen/guassian_split.dat};

	\legend{Split 1,Split 2}
\end{axis}
\end{tikzpicture}
\caption{Example of how splitting by variance fails to align with reducing bias. This split would receive a large penalty under a variance criteria, but both splits have the same mean --- resulting in no discrimination.}
\label{fig:mean_vs_var_example}
\end{figure}
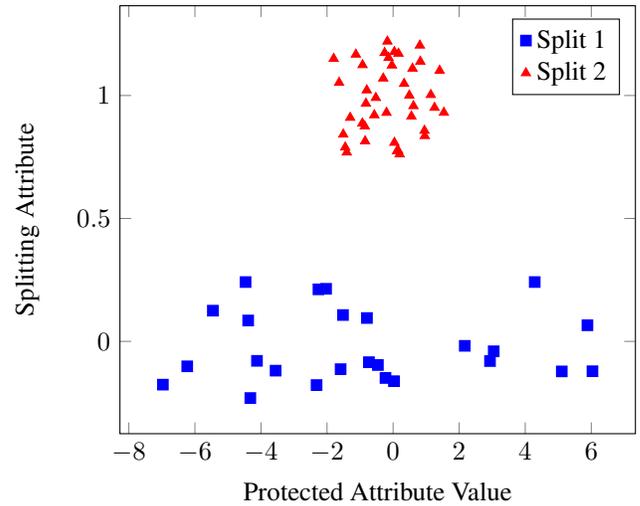

To demonstrate how this happens, consider the plot in \autoref{fig:mean_vs_var_example}. Here we show the protected attribute's value on the $x$-axis, and the attribute we are splitting on on the $y$-axis. Choosing a split of $y\geq 0.5$, we see the data cleanly splits into two groups. 

If we consider the variance-based gain defined by \eqref{eq:gain_r_variance}, this split would receive a large penalty. The split dramatically reduces the variance of the Split 2 group, which produces a large gain. Because it is the protected attribute, we subtract the gain --- and thus a large penalty is applied. 

Note though that the goal is to avoid discrimination in the predictions produced by the tree. Yet in this case, both splits have the same mean of zero, only their variances differ. If we were to think of the problem as predicting the protected attribute's value from the tree, we use the mean value of the attribute in the leaf nodes. The variance is forgotten anyway, and so we are penalizing a split which will not keep any significant information as it is. 

Thus we prefer the intuition afforded by our new gain measure \eqref{eq:gain_continuous}, which would produce no penalty for this split choice. The means would be the same, and so no predictive difference would occur with this split. Thus we find this split preferable for the protected attribute, as it would not aid in distinguishing the protected attribute at prediction time.

\end{document}